\documentclass[runningheads,a4paper]{llncs}

\usepackage{amssymb}
\usepackage{graphicx}
\usepackage{epstopdf}
\usepackage{subfigure}
\usepackage{float}
\usepackage{color}
\usepackage{makeidx}
\usepackage{amsmath}
\usepackage{multirow}

\usepackage{url}
\usepackage[table]{xcolor}
\usepackage[htt]{hyphenat}
\usepackage{array}
\newcolumntype{C}[1]{>{\centering\let\newline\\\arraybackslash\hspace{0pt}}m{#1}}
\newcolumntype{R}[1]{>{\raggedleft\let\newline\\\arraybackslash\hspace{0pt}}m{#1}}

\newcommand{\keywords}[1]{\par\addvspace\baselineskip
\noindent\keywordname\enspace\ignorespaces#1}

\begin{document}

\mainmatter  

\title{Type-Constrained Representation Learning in Knowledge Graphs}

\titlerunning{Type-Constrained Representation Learning in Knowledge Graphs}

\author{Denis Krompa\ss$^{1,2}$ \and Stephan Baier$^{1}$ \and Volker Tresp$^{1,2}$}
\authorrunning{Denis Krompa\ss \and Stephan Baier \and Volker Tresp}
\institute{
Siemens AG, Corporate Technology, Munich, Germany\\
\email{Denis.Krompass@siemens.com}\\
\email{Volker.Tresp@siemens.com}\\
Ludwig Maximilian University, 80538 Munich, Germany\\
\email{Stephan.Baier@campus.lmu.de}
}

\maketitle
\begin{abstract}
Large knowledge graphs increasingly add value to various applications that require machines to recognize and understand queries and their semantics, as in search or question answering systems.
Latent variable models have increasingly gained attention for the statistical modeling of knowledge graphs, showing promising results in tasks related to knowledge graph completion and cleaning.
Besides storing facts about the world,  schema-based knowledge graphs are backed by rich semantic descriptions of entities and relation-types that allow machines to understand the notion of things and their semantic relationships.
In this work, we study how type-constraints can generally support the statistical modeling with latent variable models.
More precisely, we integrated prior knowledge in form of  type-constraints in various state of the art latent variable approaches.
Our experimental results show that prior knowledge on relation-types significantly improves these models up to 77\% in link-prediction tasks.
The achieved improvements are especially prominent when a low model complexity is enforced, a crucial requirement when these models are applied to very large datasets.
Unfortunately, type-constraints are neither always available nor always complete e.g., they can become fuzzy when entities lack proper typing.
We  show that in these cases, it can be beneficial to apply  a local closed-world assumption that approximates the semantics of relation-types based on observations made in the data.
\end{abstract}

\section{Introduction}
Knowledge graphs (KGs), i.e., graph-based  knowledge-bases, have proven to be sources of valuable information that have become important for various applications like web-search or question answering.
Whereas, KGs were initially driven by academic efforts which resulted in KGs like Freebase\ \cite{Bollacker:2008:FCC:1376616.1376746}, DBpedia\ \cite{Bizer:2009:DCP:1640541.1640848}, Nell\ \cite{conf/aaai/CarlsonBKSHM10} or YAGO\ \cite{Hoffart:2011:YEQ:1963192.1963296}, more recently  commercial applications have evolved; a significant commercial application is the Freebase powered Google Knowledge Graph that supports Google's web search and the smart assistant Google Now,  or Microsoft's Satori that supports Bing and Cortana.
A related activity is  the linked open data initiative which  interlinks data sources using the W3C Resource Description Framework (RDF)\ \cite{Lassila1999} and thus also generates a huge KG accessible via querying\ \cite{bizer_linked_2009}.

Even though these graphs have reached an impressive size, containing billions of facts about the world, they are not error-free and far from complete.
In Freebase and DBpedia for example a vast amount of persons (71\% in Freebase\ \cite{Dong:2014:KVW:2623330.2623623} and 66\% in DBpedia) are missing a place of birth.
In DBpedia 58\% of the scientists do not have a fact that describes what they are known for.
Supporting KG cleaning, completion and construction via machine learning is one of the core challenges.
In this context, Representation Learning in form of latent variable methods has successfully been applied to KG data \cite{Nickel:2012:FYS:2187836.2187874,SocherChenManningNg2013,conf/nips/BordesUGWY13,DBLP:conf/dsaa/KrompassNT14,DBLP:conf/emnlp/ChangYYM14}.
These models learn  latent embeddings for entities and relation-types from the data that can then be used as representations of their semantics.
It is highly desirable that these embeddings are meaningful in  low dimensional latent spaces, because a higher dimensionality leads to a higher model complexities which can cause unacceptable runtime performances and high memory loads.
Latent variable models have recently been exploited for generating priors for facts in the context of automatic graph-based knowledge-base construction \cite{Dong:2014:KVW:2623330.2623623}.
It has also been shown that these models can be interpreted as a compressed probabilistic knowledge representation, which allows complex querying over all possible triples and their uncertainties, resulting in a probabilistically ranked list of query answers\ \cite{DBLP:conf/semweb/KrompassNT14}.

In addition to the stored facts, schema-based KGs also provide rich descriptions of the semantics of entities and relation-types such as class hierarchies of entities and type-constraints for relation-types which define the semantic role of relations.
This curated prior knowledge on relation-types provides valuable information to machines, e.g. that the \texttt{marriedTo} relation-type should relate only instances of the class \texttt{Person}.
In recent work \cite{DBLP:conf/dsaa/KrompassNT14,DBLP:conf/emnlp/ChangYYM14}, it has been shown that  RESCAL, a much studied latent variable approach, benefits greatly from prior knowledge about the semantics of relation-types.
In this work we will study the impact of prior knowledge about the semantics of relation-types in the state of the art representative latent variable models  TransE \cite{conf/nips/BordesUGWY13}, RESCAL\ \cite{ICML2011Nickel_438} and the multiway neural network approach used in the Google Knowledge Vault project \cite{Dong:2014:KVW:2623330.2623623}.
These models are very different in the way they model KGs, and therefore they are especially well suited for drawing conclusions on the general value of prior knowledge about relation-types for the statistical modeling of KGs with latent variable models.

Additionally, we address the issue that  type-constraints can also suffer from incompleteness, e.g.\ \texttt{rdfs:domain} or \texttt{rdfs:range} concepts are absent in the schema or the entities miss proper typing even after materialization.
Here,   we study  the local closed-world assumption as proposed in prior work \cite{DBLP:conf/dsaa/KrompassNT14} that approximates the semantics of relation-types based on observed triples.
We provide empirical proof that this prior assumption on relation-types generally improves link-prediction quality in case proper type-constraints are absent.

This paper is structured as follows:
In the next section  we motivate our model selection and briefly review RESCAL, TransE and the multiway neural network approach of \cite{Dong:2014:KVW:2623330.2623623}.
The integration of type-constraints and local closed-world assumptions into these models will be covered in Section \ref{sec:integration}.
In Section \ref{sec:experiments}, we will motivate and describe our experimental setup before we discuss our results in Section \ref{sec:results}.
We provide related work in Section \ref{sec:related} and conclude in Section \ref{sec:conclusion}.

\section{Latent Variable Models for Knowledge Graph Modeling}\label{sec:methods}
In this work, we want to study the general value of prior knowledge about the semantics of relation-types for the statistical modeling of KGs with latent variable models.
For this reason, we have to consider a representative set of latent variable models that covers the currently most promising research activities in this field.
We selected RESCAL \cite{ICML2011Nickel_438}, TransE \cite{conf/nips/BordesUGWY13} and the multiway neural network approach pursued in the Google’s Knowledge Vault project \cite{Dong:2014:KVW:2623330.2623623} (denoted as mwNN) for a number of reasons:
\begin{itemize}
  \item To the best of our knowledge, these latent variable models are the only ones which  have been applied to large KGs with more than 1 million entities, thereby proving their scalability \cite{conf/nips/BordesUGWY13,Dong:2014:KVW:2623330.2623623,Nickel:2012:FYS:2187836.2187874,DBLP:conf/emnlp/ChangYYM14,DBLP:conf/dsaa/KrompassNT14}.
  \item All of these models have been published at well respected conferences and are the basis for the most recent research activities in the field of statistical modeling of KGs (see Section \ref{sec:related}).
  \item These models are very diverse, meaning they are very different in the way they model KGs, thereby covering a wide range of possible ways a KG can be statistically modeled; the RESCAL tensor-factorization is a bilinear model, where the distance-based TransE models triples as linear translations and  the mwNN exploits non-linear interactions of latent embeddings in its neural network layers.
\end{itemize}
\subsection{Notation}
In this work, $ \mathbf{\underline{X}}$  will denote a three-way tensor, where $\mathbf{X}_k$  represents the $k$-th frontal slice of the tensor $\mathbf{\underline{X}}$.
Further $\hat{\mathbf{X}}_k$ will denote the frontal-slice $\mathbf{X}_k$ where only subject entities (rows) and object entities (columns) are included that agree with the domain and range constraints of relation-type $k$.
$\mathbf{X}$ or $\mathbf{A}$ denote  matrices and $\mathbf{x}_i$ is the $i$-th column vector of $\mathbf{X}$.
A single entry of $\mathbf{\underline{X}}$ will be denoted as $x_{i,j,k}$.
Additionally we use $\mathbf{X}_{[\mathbf{z},:]}$ to illustrate the indexing of multiple rows from the matrix $\mathbf{X}$, where $\mathbf{z}$ is a vector of indices and ``$:$'' the colon operator, generally used when indexing arrays.
Further \emph{(s,p,o)} will denote a triple with subject entity $s$, object entity $o$ and predicate relation-type $p$, where the entities $s$ and $o$ represent  nodes in the KG that are linked by the predicate relation-type $p$.
The entities belong to the set of all observed entities $\mathcal{E}$ in the data.

\subsection{RESCAL}\label{sec:rescal}
RESCAL\ \cite{ICML2011Nickel_438} is a three-way tensor factorization method that has been shown to lead to  very good results in various canonical relational learning  tasks like link-prediction, entity resolution and collective classification \cite{Nickel:2012:FYS:2187836.2187874}.
In RESCAL, triples are represented in an adjacency tensor $\mathbf{\underline{X}}$ of shape $n \times n \times m$, where $n$ is the amount of observed entities in the data and $m$ is the amount of relation-types. Each of the $m$ frontal slices $\mathbf{X}_k$ of $ \mathbf{\underline{X}}$ represents an adjacency matrix for all entities in the dataset with respect to the $k$-th relation-type.
Given an adjacency tensor $\mathbf{\underline{X}}$, RESCAL computes a rank $d$ factorization, where each entity is represented via  a $d$-dimensional vector that is stored in the factor matrix  $\mathbf{A} \in \mathbb{R}^{n\times d}$ and each relation-type is represented via a frontal slice  $\mathbf{R}_k \in \mathbb{R}^{d \times d}$ of the core tensor $\mathbf{\underline{R}}$ which encodes the asymmetric interactions between subject and object entities.
The embeddings are learned by minimizing the regularized least-squares function
\begin{equation}\label{eq:rescal}
	\mathcal{L}_{\textit{RESCAL}} = \sum_{k}^{m} \Vert \mathbf{X}_k - \mathbf{AR}_k\mathbf{A}^T \Vert^{2}_{F}
	 +\lambda_A\Vert \mathbf{A} \Vert^{2}_{F} + \lambda_R\sum_{k}^m\Vert \mathbf{R}_k \Vert^{2}_{F}~,
\end{equation}
where $\lambda_A \ge 0$ and $\lambda_R \ge 0$ are hyper-parameters and {$\| \cdot \|_F$} is the {Frobenius} norm.
The cost function can be minimized via very efficient Alternating Least-Squares (ALS) that effectively exploits data sparsity\ \cite{ICML2011Nickel_438} and closed-form solutions.
During factorization, RESCAL finds a unique latent representation for each entity that is shared between all relation-types in the dataset.

RESCAL's confidence $\theta_{s,p,o}$ for a triple $(s,p,o)$ is computed through reconstruction by the vector-matrix-vector product
\begin{equation}
\theta_{s,p,o} = \mathbf{a}_{s}^{T}\mathbf{R}_p\mathbf{a}_{o}
\end{equation}
from the latent representations of the subject and object entities $\mathbf{a}_{s}$ and $\mathbf{a}_{o}$, respectively and the latent representation of the predicate relation-type $\mathbf{R}_p$.

\subsection{Translational Embeddings Model}\label{sec:TransE}
TransE\ \cite{conf/nips/BordesUGWY13} is a distance-based model that models relationships of entities as translations in the embedding space.
The approach assumes for a true fact that a relation-type specific translation function exists that is able to map (or translate) the latent vector representation of the subject entity to the latent representation the object entity.
The  fact  confidence  is expressed by the similarity of the translation of the subject embedding to the object embedding.

In case of TransE, the translation function is defined by a simple addition of the latent vector representations of the subject entity $\mathbf{a}_s$ and the predicate relation-type $\mathbf{r}_p$.
The similarity of the translation and the  object embedding is measured by the $L_1$ or $L_2$ distance.
TransE's confidence $\theta_{s,p,o}$ in a triple $(s,p,o)$ is derived by
\begin{equation}
\theta_{s,p,o} = -\delta(\mathbf{a}_{s}+\mathbf{r}_p,\mathbf{a}_{o}),
\end{equation}
where $\delta$ is the $L_1$ or the $L2$ distance and $\mathbf{a}_{o}$ the latent embedding for the object entity.
The embeddings are learned by minimizing the max-margin-based ranking cost function
\begin{eqnarray}\label{eq:transE}
\mathcal{L}_{\textit{TransE}} &=& \sum_{(s,p,o) \in T} \text{max}\{0,\gamma + \theta_{s',p,o} - \theta_{s,p,o} \} + \text{max}\{0,\gamma + \theta_{s,p,o'} - \theta_{s,p,o} \}\notag\\
\text{with}&&~ \{s',o'\} \in \mathcal{E}
\end{eqnarray}
on a set of observed training triples $T$  through Stochastic Gradient Descent (SGD), where $\gamma >0$.
The ``corrupted'' entities $s'$ and $o'$ are drawn from the set of all observed entities $\mathcal{E}$ where the ranking loss function enforces that the  confidence in the corrupted triples ($\theta_{s',p,o}$ or $\theta_{s,p,o'}$) is lower than in the true triple by a certain margin.
During training, it is enforced that the latent embeddings of entities have an $L_2$ norm of one after each SGD iteration.

\subsection{Knowledge Vault Neural Network}\label{sec:KV}
In the Google Knowledge Vault project\ \cite{Dong:2014:KVW:2623330.2623623} a  multiway neural network (mwNN) for predicting prior probabilities for triples from existing KG data was proposed to support triple extraction from unstructured web documents.
The confidence value $\theta_{s,p,o}$ for a target triple $(s,p,o)$ is predicted by
\begin{equation}
\theta_{s,p,o} = \sigma(\mathbf{\beta}^T\phi\left(\mathbf{W}\left[\mathbf{a}_s,\mathbf{r}_p,\mathbf{a}_o\right]\right)) ,
\end{equation}
where $\phi$() is a nonlinear function like e.g.\ \emph{tanh}, $\mathbf{a}_s$ and $\mathbf{a}_o$ describe the latent embeddings for the subject and object entities and $\mathbf{r}_p$ is the latent embedding vector for the predicate relation-type $p$. $\left[\mathbf{a}_s,\mathbf{r}_p,\mathbf{a}_o\right] \in \mathbb{R}^{3d \times 1}$ is a column vector that stacks the three embeddings on top of each other. $\mathbf{W}$ and $\mathbf{\beta}$ are neural network weights and $\sigma()$ denotes the logistic  function. The model is trained by minimizing the Bernoulli cost-function
\begin{equation}\label{eq:kv}
\mathcal{L}_{\textit{mwNN}}=- \sum_{(s,p,o) \in T } \log{  \theta_{s,p,o}} -  \sum^{c}_{o' \in \mathcal{E}} \log(1-\theta_{s,p,o'})
\end{equation}
through SGD, where $c$ denotes the number of object-corrupted triples sampled under a local closed-world assumption as defined by\ \cite{Dong:2014:KVW:2623330.2623623}.
Note that corrupted are treated as negative evidence in this model.

\section{Prior Knowledge On Relation-Type Semantics}\label{sec:integration}
Generally, entities in KGs like DBpedia, Freebase or YAGO are assigned to one or multiple predefined classes (or types) that are  organized in an often hierarchical ontology.
These assignments represent for example the knowledge that the entity \texttt{Albert Einstein} is a person and therefore allow a semantic description of the entities contained in the KG.
This organization of entities in semantically meaningful classes permits a semantic definition of relation-types.
The RDF-Schema, which provides schema information  for RDF, offers among others the concepts \texttt{rdfs:domain} and \texttt{rdfs:range} for this purpose.
These concepts are used to represent type-constraints on relation-types by defining the classes or types of entities which they should relate, where the \emph{domain} covers the subject entity classes and the \emph{range} the object entity classes in a RDF-Triple.
This can be interpreted as an explicit definition of the semantics of a relation, for example by defining that the relation-type \texttt{marriedTo} should only relate instances of the class \texttt{Person} with each other.
Recently\ \cite{DBLP:conf/emnlp/ChangYYM14} and\ \cite{DBLP:conf/dsaa/KrompassNT14} showed independently that including knowledge about these domain and range constraints into RESCAL's ALS optimization scheme resulted in better latent representations of entities and relation-types that lead to a significantly improved link-prediction quality at a much lower model complexity (lower rank) when applied to KGs like DBpedia or Nell.
The need of a less complex model significantly decreases model training-time especially for larger datasets.

In the following, we denote $\mathbf{domain}_k$ as the ordered indices of all entities that agree with the domain constraints of relation-type $k$.  Accordingly, $\mathbf{range}_k$ denotes these indices for the range constraints of relation-type $k$.

\subsection{Type-Constrained Alternating Least-Squares}
In RESCAL, the integration of typed relations in the ALS optimization procedure is achieved by indexing only those latent embeddings of entities for each relation-type that agree with the \texttt{rdfs:domain} and \texttt{rdfs:range} constraints.
In addition, only the subgraph (encoded by the sparse adjacency matrix $\hat{\mathbf{X}}_k$) that is defined with respect to the constraints is considered in the equation
\begin{eqnarray}\label{eq:rescaltc}
	\mathcal{L}^{\mathcal{TC}}_{RESCAL} &=& \sum_{k} \Vert \hat{\mathbf{X}}_k - \mathbf{A}_{[\mathbf{domain}_k,:]}\mathbf{R}_k\mathbf{A}_{[\mathbf{range}_k,:]}^T\Vert^{2}_{F} \notag\\
	&&+ \lambda_A\Vert \mathbf{A} \Vert^{2}_{F} + \lambda_R\sum_{k}\Vert \mathbf{R}_k \Vert^{2}_{F},
\end{eqnarray}
where $\mathbf{A}$ contains the latent embeddings for  the entities and $\mathbf{R}_k$ the embedding for the relation-type $k$.
For each relation-type $k$ the latent embeddings matrix $\mathbf{A}$ is indexed by the corresponding domain and range constraints, thereby excluding all entities that disagree with the type-constraints.
Note that if the adjacency matrix $\hat{\mathbf{X}}_k$ of the subgraph defined by relation-type $k$ and its type-constraints has the shape $n_k \times m_k$, then $\mathbf{A}_{[\mathbf{domain}_k,:]}$ is of shape $n_k \times d$, and $\mathbf{A}_{[\mathbf{range_k},:]}$ of shape $m_k \times d$ where $d$ is the dimension of the latent embeddings (or rank of the factorization).

\subsection{Type-Constrained Stochastic Gradient Descent}
In contrast to RESCAL, TransE and mwNN are both optimized through mini-batch Stochastic Gradient Descent (SGD), where a small batch of randomly sampled triples is used in each iteration of the optimization to drive the model parameters to a local minimum.
Generally, KG data does not explicitly contain negative evidence, i.e.\ false triples \footnote{There are of course undetected false triples included in graph which are assumed to be true.}, and is generated in this algorithms through corruption of observed triples (see Section \ref{sec:TransE} and \ref{sec:KV}).
In the original algorithms of  TransE and mwNN the corruption of triples is not restricted and can therefore lead to the generation of triples that violate the semantics of relation-types.
For integrating knowledge about type-constraints into the SGD optimization scheme of these models, we have to make sure that none of the corrupted triples violates the type-constraints of the corresponding relation-types.
For TransE we update Equation \ref{eq:transE} and get
\begin{eqnarray}\label{eq:transEtc}
\mathcal{L}^{\mathcal{TC}}_{TransE} &=& \sum_{(s,p,o) \in T}\sum_{(s',p,o') \in T'} \left[\gamma + \theta_{s',p,o} - \theta_{s,p,o} \right]_{+} + \left[\gamma + \theta_{s,p,o'} - \theta_{s,p,o} \right]_{+}\notag\\
\text{with}&&~ s'\in\mathcal{E}_{[\mathbf{domain}_p]}\subseteq\mathcal{E},~o'\in\mathcal{E}_{[\mathbf{range}_p]}\subseteq\mathcal{E},
\end{eqnarray}
where, in difference to Equation \ref{eq:transE}, we enforce by  $s'\in\mathcal{E}_{[\mathbf{domain}_p]}\subseteq\mathcal{E}$ that the subject entities are only corrupted through the subset of entities that belong to the domain and  by  $o'\in\mathcal{E}_{[\mathbf{range}_p]}\subseteq\mathcal{E}$ that the corrupted object entities are sampled  from the subset of entities that belong to the range of predicate relation-type $p$.
For mwNN we corrupt only the object entities through sampling from the subset of entities $o'\in\mathcal{E}_{[\mathbf{range}_p]}\subseteq\mathcal{E}$ that belong to the range of the predicate relation-type $p$ and get accordingly
\begin{equation}\label{eq:kvtc}
\mathcal{L}^{\mathcal{TC}}_{\textit{mwNN}}=-\sum_{(s,p,o) \in T } \log{  \theta_{s,p,o}} -  \sum^{c}_{o'\in\mathcal{E}_{[\mathbf{range}_p]}\subseteq\mathcal{E}} \log(1-\theta_{s,p,o'}) .
\end{equation}

\subsection{Local Closed-World Assumptions}\label{sec:lcwa}
Type-constraints as given by KGs tremendously reduce the possible worlds of the statistically modeled KGs, but like the rest of the data represented by the KG, they can also suffer from incompleteness and inconsistency of the data.
Even after materialization,  entities and relation-types might miss complete typing leading to fuzzy type-constraints.
Increased fuzziness of proper typing can in turn lead to disagreements of true facts and present type-constraints in the KG.
For relation-types where these kind of inconsistencies are quite frequent  we cannot simply apply the given type-constraints without the risk of loosing true triples.
On the other hand, if the domain and range constraints themselves are missing (e.g.\ in schema-less KGs) we might consider many triples that do not have any semantic meaning.

We argue that in these cases a local closed-world assumption (LCWA) can be applied which approximates the domain and range constraints of the targeted relation-type not on class level, but on instance level based solely on observed triples.
Given all observed triples, under this LCWA the domain of a relation-type $k$ consists of all entities that are related by the relation-type $k$ as subject.
The range is accordingly defined, but contains all the entities related as object by relation-type $k$.
Of course, this approach can exclude entities from the domain or range constraints that agree with the type-constraints given by the RDFS-Schema concepts \texttt{rdfs:domain} and \texttt{rdfs:range}, thereby ignoring them during model training when exploiting the LCWA (only for the target relation-type).
On the other hand, nothing is known about these entities (in object or subject role) with respect to the target relation-type and therefore treating them as missing can be a valid assumption.
In case of the ALS optimized RESCAL we reduce the size and sparsity of the data by this approach, which has a positive effect on model training compared to the alternative, a closed-world assumption that considers all entities to be part of the domain and range of the target relation-type\ \cite{DBLP:conf/dsaa/KrompassNT14}.
For the SGD optimized TransE and mwNN models also a positive effect on the learned factors is expected since the corruption of triples will be based on entities from which we can expect that they do not disagree to the semantics of the corresponding relation-type.

\section{Experimental Setup}\footnote{Code and datasets will be available from http://www.dbs.ifi.lmu.de/$\sim$krompass/}\label{sec:experiments}
As stated before, we explore in our experiments  the importance of prior knowledge about the semantics of relation-types for latent variable models.
We consider two settings.
In the first setting, we assume that  curated type-constraints extracted from the KG's schema are available.
In the second setting, we explore the  local closed-world assumption (see Section \ref{sec:lcwa}).
Our experimental setup covers three important aspects which will enable us to make generalizing conclusions about the importance of such prior knowledge when applying latent variable models to KGs:
\begin{itemize}
\item We test various representative latent variable models that cover the diversity of these models in the domain.
As motivated in the introduction of Section \ref{sec:methods},  we belief that  RESCAL, TransE and mwNN are especially well suited for this task.
\item We test these models at reasonable low complexity levels, meaning that we enforce low dimensional latent embeddings, which simulates their application to very large datasets where high dimensional embeddings are intractable.
In\ \cite{Dong:2014:KVW:2623330.2623623} for example, a latent embedding length  $d=60$  (see Section~\ref{sec:KV})  was used.
\item We extracted diverse datasets from instances of the Linked-Open Data Cloud, namely Freebase, YAGO and DBpedia, because it is  expected that the value of prior knowledge about relation-type semantics is also dependent on the particular dataset the models are applied to.
From these KGs we constructed datasets that will be used as representatives for  general purpose KGs that cover a wide range of relation-types from a diverse set of domains, domain focused KGs with a small amount of entity classes  and relation-types and high quality KGs.
\end{itemize}
In the remainder of this section we will give details on the extracted datasets and the evaluation, implementation and training of RESCAL, TransE and mwNN.

\subsection{Datasets}\label{sec:Experiments:Datasets}
\begin{table}[t]
 \caption{Datasets used in the experiments.}
 \centering
\begin{tabular}{l|l|r|r|r}
  \hline
  \textbf{Dataset} & \textbf{Source} & \textbf{Entities}  & \textbf{Relation-Types} & \textbf{Triples} \\ \hline
  DBpedia-Music  & DBpedia 2014      &   321,950 &         15 &    981,383 \\
  Freebase-150k  & Freebase RDF-Dump &   151,146 &        285 &  1,047,844 \\
  YAGOc-195k     & YAGO2-Core        &   195,639 &         32 &  1,343,684 \\
\end{tabular}
\label{tab:datasets}
\end{table}
Below, we describe how we extracted the different datasets from Freebase, DBpedia and YAGO.
In Table \ref{tab:datasets} some details about the size of these datasets are given.
In our experiments, the Freebase-150k dataset will simulate a general purpose KG, the DBpedia-Music dataset a domain specific KG and the YAGOc-195k dataset a high quality KG. 

\subsubsection{Freebase-150k}
The Freebase KG includes triples extracted from Wikipedia Infoboxes, MusicBrainz\ \cite{journals/expert/Swartz02}, WordNet\ \cite{Miller:1995:WLD:219717.219748} and many more.
From the current materialized Freebase RDF-dump\footnote{https://developers.google.com/freebase/data}, we extracted entity-types, type-constraints and all triples that involved entities (\emph{Topics}) with more than  100 relations to other topics.
Subsequently, we discarded the triples of relation-types with incomplete type-constraints or which occurred in less than 100 triples.
Additionally,  we discarded all triples that involved entities that are not an instance of any class covered by the remaining type-constraints.
The entities involved in type-constraint violating triples were added to the subset of entities that agree with the type-constraints since we assumed that they only miss proper typing.

\subsubsection{DBpedia-Music}
For the DBpedia-Music datasets, we extracted triples and types from 15 pre-selected object-properties regarding the music domain of DBpedia \footnote{http://wiki.dbpedia.org/Downloads2014, canonicalized datasets: mapping-based-properties(cleaned), mapping-based-types and heuristics}; \texttt{musicalBand}, \texttt{musicalArtist}, \texttt{musicBy}, \texttt{musicSubgenre}, \texttt{derivative}, \texttt{stylisticOrigin}, \texttt{associatedBand}, \texttt{associatedMusicalArtist}, \texttt{recordedIn}, \texttt{musicFusionGenre}, \texttt{musicComposer}, \texttt{artist}, \texttt{bandMember}, \texttt{formerBandMember}, \texttt{genre}, where  \texttt{genre} has been extracted to include only those entities that were covered by the other
object-properties to restrict it to musical genres.
We extracted the type-constraints from the DBpedia OWL-Ontology and for entities that occurred less than two times we discarded all triples.
In case types for entities or type-constraints were absent we assigned them to \texttt{owl\#Thing}.
Remaining disagreements between triples and type-constraints were resolved as in case of the Freebase-150k dataset.

\subsubsection{YAGOc-195k}
YAGO (Yet Another Great Ontology) is an automatically generated high quality KG that combines the information richness of Wikipedia Infoboxes and its category system with the clean taxonomy of WordNet.
We extracted entitiy types, type-constraints\footnote{yagoSchema and yagoTransitiveType} and all triples that involved entities with more than 5 and relation-types that were involved in more than 100 relations from the YAGO-core dataset\footnote{http://www.mpi-inf.mpg.de/departments/databases-and-information-systems/research/yago-naga/yago/downloads/}.
We only included entities that share the types used in the \texttt{rdfs:domain} and \texttt{rdfs:range} triples.

\subsection{Evaluation Procedure}\label{sec:Experiments:Training}
We evaluate RESCAL, TransE and mwNN on link prediction tasks, where we delete triples from the datasets and try to re-predict them without considering them during model training.
For model training and evaluation we split the triples of the datasets into three sets, where 20\% of the triples were taken as holdout set, 10\% as validation set for hyper-parameter tuning and the remaining 70\% served as training set\footnote{additional 5\% of the training set were used for early stopping}.
In case of the validation and holdout set, we sampled 10 times as many negative triples for evaluation, where the negative triples were drawn such that they did not violate the given domain and range constraints of the KG.
Also, the negative evidence of the holdout and validation set are not overlapping.
In KG data, we are generally dealing with a strongly skewed ratio of observed and unobserved triples, through this sampling we try to mimic this effect to some extend since it is intractable to sample all unobserved triples.
In case of the LCWA, the domain and range constraints are always derived from the training set.
After deriving the best hyper-parameter settings for all models, we trained all models with these settings using both, the training and the validation set to predict the holdout set (20\% of triples).
We report the Area Under Precision Recall Curve (AUPRC) for all models.
In addition, we provide the Area Under Receiver Operating Characteristic Curve (AUROC), because it is widely used in this problem even though it is not well suited for evaluation in these tasks due to the imbalance of (assumed) false and true triples.\footnote{AUROC considers the false-positive rate which relies on the amount of true-negatives that is generally high in these kind of datasets resulting in misleadingly high scores.}
The discussions and conclusions will be primarily based on the AUPRC results.

\subsection{Implementation and Model Training Details}
All models were implemented in Python using in part Theano\ \cite{bergstra-proc-scipy-2010}.
For TransE we exploited the code provided by the authors \footnote{https://github.com/glorotxa/SME} as a basis to implement a type-constraints supporting version of TransE, but
we replaced large parts of the original code to allow a significantly faster training.\footnote{Mainly caused by the ranking function used for calculating the validation error but also the consideration of trivial zero gradients during the SGD-updates.}
We made sure that our implementation achieved very similar results to the original model on a smaller dataset\footnote{http://alchemy.cs.washington.edu/data/cora/} (results not shown).

The mwNN was also implemented in Theano.
Since there are not many details on model training in the corresponding work\ \cite{Dong:2014:KVW:2623330.2623623}, we  added elastic-net regularization combined with DropConnect\ \cite{conf/icml/WanZZLF13} on the network weights and optimized the cost function using mini-batch adaptive gradient descent.
We randomly initialized the weights by drawing from a zero mean normal distribution where we treat the standard deviation as an additional hyper-parameter.
The corrupted triples were sampled with respect to the local closed-world assumption discussed in\ \cite{Dong:2014:KVW:2623330.2623623}.
We fixed the amount of corrupted triples per training example to five.\footnote{We tried different amounts of corrupted triples and five seemed to give the most stable results across all datasets}

For RESCAL, we used the ALS implementation provided by the author\footnote{https://github.com/mnick/scikit-tensor} and our own implementation used in\  \cite{DBLP:conf/dsaa/KrompassNT14}, but modified them such that they support a more scalable early stopping criteria based on a small validation set.

For hyper-parameter tuning, all models were trained for a maximum of 50 epochs and for the final evaluation on the holdout set for a maximum of 200 epochs.
For all models, we sampled 5\% of the training data and used the change in AUPRC on this subsample as early stopping criteria.

\section{Experimental Results}\label{sec:results}
In tables \ref{tab:rescal}, \ref{tab:transE} and \ref{tab:kvnn} our experimental results for RESCAL, TransE and mwNN are shown.
All of these tables have the same structure and compare different versions of exactly one of these methods on all three datasets.
Table \ref{tab:rescal} for example shows the results for RESCAL and Table \ref{tab:kvnn} the results of mwNN.
The first column in these tables indicates the datasets the model was applied to (Freebase-150k, Dbpedia-Music or YAGOc-195) and the second column which kind of prior knowledge about the semantics of relation-types was exploited by the model.
\emph{None} denotes in this case the original model that does not consider any prior knowledge on relation-types, whereas \emph{Type-Constraints} denotes that the model has exploited the curated domain and range constraints extracted from the KG's schema and \emph{LCWA} that the model has exploited the Local Closed-World Assumption (Section \ref{sec:lcwa}) during model training.
The last two columns show the AUPRC and AUROC scores for the various model versions on the different datasets.
Each of these two columns contains three sub-columns that show the AUPRC and AUROC scores at different enforced latent embedding lengths: 10, 50 or 100.
\begin{table}[ht]
 \caption{Comparison of AUPRC and AUROC result for RESCAL with and without exploiting prior knowledge about relations types (type-constraints or local closed-world assumption (LCWA)) on the Freebase, DBpedia and YAGO2 datasets. \emph{d} is representative for the model complexity, denoting the enforced length of the latent embeddings (rank of the factorization). }
 \centering
\begin{tabular}{l|l|R{1cm}R{1cm}R{1cm}|R{1cm}R{1cm}R{1cm}}
 \cellcolor{black!25} 
& \multirow{2}{2.8cm}{\textbf{Prior Knowledge on Semantics}} & \multicolumn{3}{c}{\textbf{AUPRC}}&\multicolumn{3}{|c}{\textbf{AUROC}}\\
\multicolumn{1}{c|}{\multirow{-2}{*}{\cellcolor{black!25}
\textbf{\underline{RESCAL}}}} & & \textbf{\emph{d}}=10 & \textbf{\emph{d}}=50  & \textbf{\emph{d}}=100   & \textbf{\emph{d}}=10  & \textbf{\emph{d}}=50  & \textbf{\emph{d}}=100   \\
\hline
\multirow{3}{*}{\textbf{Freebase-150k}} & None		      	&         0.327 &         0.453 &         0.514 &         0.616 &         0.700 &         0.753\\
                                        & Type-Constraints	&         0.521 &         0.630 &         0.654 &         0.804 &         0.863 &         0.877\\
                                        & LCWA				&\textbf{0.579} &\textbf{0.675} &\textbf{0.699} &\textbf{0.849} &\textbf{0.886} &\textbf{0.896}\\
\hline
\noalign{\global\arrayrulewidth0.4pt}
\multirow{3}{*}{\textbf{DBpedia-Music}} & None             &         0.307 &         0.362 &         0.416 &         0.583 &         0.617 &         0.653\\
                                        & Type-Constraints &         0.413 &         0.490 &         0.545 &         0.656 &         0.732 &         0.755\\
                                        & LCWA             &\textbf{0.453} &\textbf{0.505} &\textbf{0.571} &\textbf{0.701} &\textbf{0.776} &\textbf{0.800}\\
\hline
\noalign{\global\arrayrulewidth0.4pt}
\multirow{3}{*}{\textbf{YAGOc-195k}}    & None             &         0.507 &         0.694 &         0.721 &         0.621 &         0.787 &         0.800\\
                                        & Type-Constraints &\textbf{0.626} &\textbf{0.721} &\textbf{0.739} &         0.785 &         0.820 &         0.833\\
                                        & LCWA             &         0.567 &         0.672 &         0.680 &\textbf{0.814} &\textbf{0.839} &\textbf{0.849}\\
  \end{tabular}
\label{tab:rescal}
\end{table}
\begin{table}[ht]
 \caption{Comparison of AUPRC and AUROC result for TransE with and without exploiting prior knowledge about relations types (type-constraints or local closed-world assumption (LCWA)) on the Freebase, DBpedia and YAGO2 datasets. \emph{d} is representative for the model complexity, denoting the enforced length of the latent embeddings.}
 \centering
\begin{tabular}{l|l|R{1cm}R{1cm}R{1cm}|R{1cm}R{1cm}R{1cm}}
\cellcolor{black!25} 
& \multirow{2}{2.8cm}{\textbf{Prior Knowledge on Semantics}} & \multicolumn{3}{c}{\textbf{AUPRC}}&\multicolumn{3}{|c}{\textbf{AUROC}}\\
\multicolumn{1}{c|}{\multirow{-2}{*}
{
\cellcolor{black!25}
\textbf{\underline{TransE}}}} & & \textbf{\emph{d}}=10 & \textbf{\emph{d}}=50  & \textbf{\emph{d}}=100   & \textbf{\emph{d}}=10  & \textbf{\emph{d}}=50  & \textbf{\emph{d}}=100   \\
\hline
\multirow{3}{*}{\textbf{Freebase-150k}} & None			&         0.548 &         0.715 &         0.743 &         0.886 &         0.890 &         0.892\\
								  & Type-Constraints	&\textbf{0.699} &         0.797 &         0.808 &\textbf{0.897} &         0.918 &         0.907\\
								  & LCWA				&         0.671 &\textbf{0.806} &\textbf{0.831} &         0.894 &\textbf{0.932} &\textbf{0.931}\\
\hline
\noalign{\global\arrayrulewidth0.4pt}
\multirow{3}{*}{\textbf{DBpedia-Music}} & None             &         0.701 &         0.748 &         0.745 &         0.902 &         0.911 &         0.903\\
                                        & Type-Constraints &\textbf{0.734} &         0.783 &         0.826 &\textbf{0.927} &         0.937 &         0.942\\
                                        & LCWA             &         0.719 &\textbf{0.839} &\textbf{0.848} &         0.910 &\textbf{0.943} &\textbf{0.953}\\
\hline
\noalign{\global\arrayrulewidth0.4pt}
\multirow{3}{*}{\textbf{YAGOc-195}}    & None             &         0.793 &         0.849 &         0.816 &         0.904 &         0.960 &         0.910\\
                                        & Type-Constraints &\textbf{0.843} &\textbf{0.896} &\textbf{0.896} &\textbf{0.962} &\textbf{0.972} &\textbf{0.974} \\
                                        & LCWA             &         0.790 &         0.861 &         0.872 &         0.942 &         0.962 &         0.962\\
  \end{tabular}
\label{tab:transE}
\end{table}
\begin{table}[ht]
 \caption{Comparison of AUPRC and AUROC result for mwNN\ \cite{Dong:2014:KVW:2623330.2623623} with and without exploiting prior knowledge about relations types (type-constraints or local closed-world assumption (LCWA)) on the Freebase, DBpedia and YAGO2 datasets. \emph{d} is representative for the model complexity, denoting the enforced length of the latent embeddings. }
 \centering
\begin{tabular}{l|l|R{1cm}R{1cm}R{1cm}|R{1cm}R{1cm}R{1cm}}
\cellcolor{black!25} 
& \multirow{2}{2.8cm}{\textbf{Prior Knowledge on Semantics}} & \multicolumn{3}{c}{\textbf{AUPRC}}&\multicolumn{3}{|c}{\textbf{AUROC}}\\
\multicolumn{1}{c|}{\multirow{-2}{*}{
\cellcolor{black!25}
\textbf{\underline{mwNN}}}} & & \textbf{\emph{d}}=10 & \textbf{\emph{d}}=50  & \textbf{\emph{d}}=100   & \textbf{\emph{d}}=10  & \textbf{\emph{d}}=50  & \textbf{\emph{d}}=100   \\
\hline
\multirow{3}{*}{\textbf{Freebase-150k}} & None			&         0.437 &         0.471 &         0.512 &         0.852 &         0.868 &         0.879\\
								  & Type-Constraints	&\textbf{0.775} &\textbf{0.815} &\textbf{0.837} &\textbf{0.956} &\textbf{0.962} &\textbf{0.967}\\
								  & LCWA				&         0.610 &         0.765 &         0.776 &         0.918 &         0.954 &         0.956\\
\hline
\noalign{\global\arrayrulewidth0.4pt}
\multirow{3}{*}{\textbf{DBpedia-Music}} & None             &         0.436 &         0.509 &         0.538 &         0.836 &         0.864 &         0.865\\
                                        & Type-Constraints &         0.509 &\textbf{0.745} &\textbf{0.754} &         0.858 &\textbf{0.908} &\textbf{0.913}\\
                                        & LCWA             &\textbf{0.673} &         0.707 &         0.723 &\textbf{0.876} &         0.900 &         0.884\\
\hline
\noalign{\global\arrayrulewidth0.4pt}
\multirow{3}{*}{\textbf{YAGOc-195}}    & None             &         0.600 &         0.684 &         0.655 &         0.949 &         0.949 &         0.957\\
                                        & Type-Constraints &\textbf{0.836} &\textbf{0.840} &\textbf{0.837} &\textbf{0.953} &\textbf{0.954} &\textbf{0.960}\\
                                        & LCWA             &         0.714 &         0.836 &         0.833 &         0.926 &         0.935 &         0.943\\
  \end{tabular}
\label{tab:kvnn}
\end{table}
\subsection{Type-Constraints are Essential}\label{sec:tc_results}
The experimental results shown in Table \ref{tab:rescal}, \ref{tab:transE} and \ref{tab:kvnn} give strong evidence that type-constraints as provided by the KG's schema are generally of great value for the statistical modeling of KGs with latent variable models.
For all datasets, this prior information lead to significant improvements in link-prediction quality for all models and settings in both, AUPRC and AUROC.
For example, RESCAL's, AUPRC score on the Freebase-150k dataset gets improved from 0.327 to 0.521 at the lowest model complexity ($d=10$) (Table \ref{tab:rescal}).
With higher model complexities the relative improvements decrease but stay significant (27\% at $d=100$ from 0.514 to 0.654).
The benefit for RESCAL in considering  type-constraints was expected due to prior works\ \cite{DBLP:conf/emnlp/ChangYYM14,DBLP:conf/dsaa/KrompassNT14}, but also the other models improve significantly when considering type-constraints.

For TransE, large improvements on the Freebase-150k and DBpedia-Music datasets can be observed (Table \ref{tab:transE}), where the AUPRC score increases e.g.\ for $d=10$ from 0.548 to 0.699 in Freebase-150k and for  $d=100$ from 0.745 to 0.826 in DBpedia-Music.
Also on the YAGOc-195k dataset the link-prediction quality improves from 0.793 to 0.843 with $d=10$.
Especially the multiway neural network approach (mwNN) seems to improve the most by considering type-constraints during the model training (Table \ref{tab:kvnn}).
In case of the Freebase-150k dataset, it improves up to 77\% in AUPRC for $d=10$ from 0.437 to 0.775 and on the DBpedia-Music dataset from  0.436 to 0.509 with $d=10$ and from 0.538 to 0.754  with $d=100$ in AUPRC.
In case of the YAGOc-195k dataset the link-prediction quality of mwNN also benefits to a large extent from the type-constraints.

Besides observing that the latent variable models are superior when  exploiting type-constraints at a fixed latent embedding length $d$, it is also worth noticing that the biggest improvements are most often achieved at a very low model complexity ($d=10$), which is especially interesting for the application of these models to large datasets.
At this low complexity level the type-constraints supported models even outperform more complex counterparts that ignore type-constraints, e.g.\ on Freebase-150k mwNN reaches 0.512 AUPRC with an embedding length of 100  but by considering type-constraints this models achieves 0.775 AUPRC with an embedding length of only 10.

In accordance to the AUPRC scores, the improvements of the less meaningful and generally high AUROC scores support the conclusion that type-constraints add value to the prediction quality of the models.
It can be inferred from the corresponding scores that the improvements have a smaller scale, but are still significant.

\subsection{Local Closed-World Assumption -- Simple but Powerful}
From Tables \ref{tab:rescal}, \ref{tab:transE} and \ref{tab:kvnn}, it can be observed that the LCWA leads to similar large improvements in link-prediction quality than the real type-constraints, especially at the lowest model complexities ($d=10$).
For example, by exploiting the LCWA TransE improves from 0.715 to 0.806 with $d=50$ in the Freebase-150k dataset, mwNN improves its initial AUPRC score of 0.600 ($d=10$) on the YAGO dataset to 0.714 and  RESCAL's AUPRC score jumps from 0.327 to 0.579 ($d=10$).
The only exception to this observation is RESCAL when applied to the YAGOc-195k dataset.
For $d=50$,  the RESCAL AUPRC score decreases from 0.694 to 0.672 and for $d=100$ from 0.721 to 0.680 AUPRC when considering the LCWA in the model.
The type-constraints of the YAGOc-195k relation-types are defined over a large set of entities, covering 22\% of all possible triples
It seems that a closed-world assumption is more beneficial for RESCAL in this case.
As in case of the type-cnstraints, the AUROC scores also support the trend observed through the AUPRC scores.

Even though the LCWA has a similar beneficial impact on the link-prediction quality than the type-constraints, there is no evidence in our experiments that the LCWA can generally replace the extracted type-constraints provided by the KG's schema.
For the YAGOc-195k dataset, the type-constraint supported models are clearly superior to those that exploit the LCWA, but in case of the Freebase-150k and DBpedia-Music datasets the message is not as clear.
RESCAL achieves on these two datasets its best results when exploiting LCWA where mwNN achieves its best results when exploiting the type-constraints.
For TransE it seems to depend on the chosen embedding length, where longer embedding lengths favor the LCWA.


\section{Related Work}\label{sec:related}
A number of other latent variable models have been proposed for the statistical modeling of KGs.
\cite{SocherChenManningNg2013} recently proposed a neural tensor network, which we did not consider in our study, since it was observed that it does not scale to larger datasets\ \cite{DBLP:conf/emnlp/ChangYYM14,Dong:2014:KVW:2623330.2623623}.
Instead we exploit a less complex and more scalable neural network model proposed in\ \cite{Dong:2014:KVW:2623330.2623623}, which could achieve comparable results to the neural tensor network of\ \cite{SocherChenManningNg2013}.
TransE\ \cite{conf/nips/BordesUGWY13} has been target of other recent research activities.
\cite{DBLP:journals/corr/YangYHGD14a} proposed a framework for relationship modeling that combines aspects of TransE and the neural tensor network proposed in\  \cite{SocherChenManningNg2013}.
\cite{DBLP:conf/aaai/WangZFC14} proposed TransH which improves TransE's capability to model reflexive one-to-many, many-to-one and many-to-many relation-types by introducing a relation-type specific hyperplane where the translation is performed.
This work has been further extended in\ \cite{DBLP:conf/aaai/LinLSLZ15} by introducing TransR which separates representations of entities and relation-types in different spaces, where the translation is performed in the relation-space.
An extensive review on representation learning with KGs can be found in\ \cite{DBLP:journals/corr/NickelMTG15}.

Domain and range constraints as given by the KG's schema or via a local closed-world assumption have been exploited very recently in RESCAL\ \cite{DBLP:conf/emnlp/ChangYYM14,DBLP:conf/dsaa/KrompassNT14}, but to the best of our knowledge have not yet been integrated into other latent variable methods nor has their general value been recognized for these models.

Further, latent variable methods have been combined with graph-feature models which lead to an increase of prediction quality\ \cite{Dong:2014:KVW:2623330.2623623} and a decrease of model complexity\ \cite{DBLP:conf/nips/NickelJT14}.

\section{Conclusions and Future Work}\label{sec:conclusion}
In this work we have studied the general value of prior knowledge about the semantics of relation-types, extracted from the schema of the knowledge graph (type-constraints) or approximated through a local closed-world assumption,  for the statistical modeling of KGs with latent variable models.
Our experiments give clear empirical proof that the curated semantic information of type-constraints significantly improves link-prediction quality of TransE, RESCAL and mwNN (up to 77\%) and can therefore be considered as essential for latent variable models when applied to KGs.
Thereby the value of type-constraints becomes especially prominent when the model complexity, i.e.\ the dimensionality of the embeddings has to be very low, an essential requirement when applying these models to very large datasets.

Since type-constraints can be absent or fuzzy (due to e.g.\ insufficient typing of entities), we further showed that an alternative, a local closed-world assumption (LCWA), can be applied in these cases that approximates domain range constraints for relation-types on instance level rather on class level solely based on observed triples.
This LCWA also leads to large improvements in the link-prediction tasks, but especially at a very low model complexity the integration of type-constraints seemed superior.
In our experiments we used models that either exploited type-constraints or the LCWA, but in a real setting we would combine both, where we would use the type-constraints whenever possible, but the LCWA on the relation-types where type-constraints are absent or fuzzy.

In future work we will further investigate on additional extensions for latent variable models that can be combined with the type-constraints or LCWA. In the related-work we gave some examples were the integration of graph-feature models (e.g.\ the path ranking algorithm\ \cite{Lao:2010:RRU:1842816.1842823}) was shown to improve these models.
In addition we will look at the many aspects in which RESCAL, TransE and mwNN differ.
Identifying the aspects of these models that have the most beneficial impact on link-prediction quality can give rise to a new generation of latent variable approaches that could further drive knowledge graph modeling.

\bibliographystyle{plain}
\bibliography{references_backup}

\end{document}